\documentclass[10pt,twocolumn,letterpaper]{article}

\usepackage{cvpr}              % To produce the CAMERA-READY version
\usepackage{graphicx}
\usepackage{amsmath}
\usepackage{amssymb}
\usepackage{booktabs}
\usepackage[dvipsnames]{xcolor}
\usepackage{caption}
\usepackage{subcaption}

\usepackage[pagebackref,breaklinks,colorlinks]{hyperref}

\usepackage[capitalize]{cleveref}
\crefname{section}{Sec.}{Secs.}
\Crefname{section}{Section}{Sections}
\Crefname{table}{Table}{Tables}
\crefname{table}{Tab.}{Tabs.}

\def\cvprPaperID{06} % *** Enter the CVPR Paper ID here
\def\confName{CVPR}
\def\confYear{2022}

\newcommand{\token}[1]{\textup{[}\textit{#1}\textup{]}}
\begin{document}

%%%%%%%%% TITLE - PLEASE UPDATE
\title{VL-InterpreT: An Interactive Visualization Tool for Interpreting Vision-Language Transformers}

\author{Estelle Aflalo\footnotemark[1]\\
Intel Labs\\
{\tt\small estelle.aflalo@intel.com}
\and
Meng Du\footnotemark[1]\\
Intel Labs, UCLA\\
{\tt\small mengdu@ucla.edu}
\and
Shao-Yen Tseng\\
Intel Labs\\
{\tt\small shao-yen.tseng@intel.com}
\and
Yongfei Liu\\
Microsoft Research\\
{\tt\small liuyf3@shanghaitech.edu.cn}
\and
Chenfei Wu\\
Microsoft Research\\
{\tt\small chewu@microsoft.com}
\and
Nan Duan\\
Microsoft Research\\
{\tt\small nanduan@microsoft.com}
\and
Vasudev Lal\\
Intel Labs\\
{\tt\small vasudev.lal@intel.com}
}

\maketitle

\renewcommand*{\thefootnote}{\fnsymbol{footnote}}
\footnotetext[1]{Equal Contributions}
\renewcommand*{\thefootnote}{\arabic{footnote}}

%%%%%%%%% ABSTRACT
\begin{abstract}
Breakthroughs in transformer-based models have revolutionized not only the NLP field, but also vision and multimodal systems. However, although visualization and interpretability tools have become available for NLP models, internal mechanisms of vision and multimodal transformers remain largely opaque. With the success of these transformers, it is increasingly critical to understand their inner workings, as unraveling these black-boxes will lead to more capable and trustworthy models. To contribute to this quest, we propose VL-InterpreT, which provides novel interactive visualizations for interpreting the attentions and hidden representations in multimodal transformers. VL-InterpreT is a task agnostic and integrated tool that (1) tracks a variety of statistics in attention heads throughout all layers for both vision and language components, (2) visualizes cross-modal and intra-modal attentions through easily readable heatmaps, and (3) plots the hidden representations of vision and language tokens as they pass through the transformer layers. In this paper, we demonstrate the functionalities of VL-InterpreT through the analysis of KD-VLP, an end-to-end pretraining vision-language multimodal transformer-based model, in the tasks of Visual Commonsense Reasoning (VCR) and WebQA, two visual question answering benchmarks. Furthermore, we also present a few interesting findings about multimodal transformer behaviors that were learned through our tool. 

\end{abstract}

%%%%%%%%% BODY TEXT
\section{Introduction}
Since transformers were introduced in Vaswani \etal \cite{NIPS2017_3f5ee243}, not only have they seen massive success in NLP applications, their impact on computer vision and multimodal problems has also become increasingly disruptive. However, the internal mechanisms of transformers that lead to such successes are not well understood. Although efforts have been made to interpret the attentions~\cite{Clark2019WhatDB} and hidden states~\cite{liu-etal-2019-linguistic} of transformers for NLP, such as BERT~\cite{devlin-etal-2019-bert}, investigations in the mechanisms of vision and multimodal transformers are relatively scarce, and tools for probing such transformers are also limited. Given the fast-growing number of successful vision and multimodal transformers (\eg, ViT~\cite{dosovitskiy2021an} and CLIP~\cite{Radford2021LearningTV}), enhanced interpretability of these models is needed to guide better designs in the future.

Past research has shown the importance of interpreting the inner mechanisms of transformers. For example, Clark \etal~\cite{Clark2019WhatDB} found certain BERT attention heads specialized in handling certain syntactic relations, as well as interesting ways in which BERT attention utilizes special tokens and punctuation marks. Additionally, Lin \etal \cite{lin-etal-2019-open} showed that the linguistic information encoded in BERT becomes increasingly abstract and hierarchical in later layers. These studies provide valuable insights into the functions of various elements in transformer architecture for NLP, and shed light on their limitations.

This paper presents VL-InterpreT\footnote{A screencast of our application is available at \url{https://www.youtube.com/watch?v=4Rj15Hi_Pdo}. Source code and a link to a live demo: \url{https://github.com/IntelLabs/VL-InterpreT}}, which is an interactive visualization tool for interpreting the attentions and hidden representations of vision-language (VL) transformers. Importantly, it is a single system that analyzes and visualizes several aspects of multimodal transformers: first, it tracks behaviors of both vision and language attention components in attention heads throughout all layers, as well as the interactions across the two modalities. Second, it also visualizes the hidden representations of vision and language tokens as they pass through transformer layers.

The main contributions of our work are:
\begin{itemize}
\item Our tool allows interactive visualization for probing hidden representations of tokens in VL transformers.

\item Our tool allows systematic analysis, interpretation, and interactive visualization of cross- and intra-modal components of attention in VL transformers.

\item As an application of VL-InterpreT, we demonstrate multimodal coreference in two analyses: 1) how contextualized tokens in different modalities referring to the same concept are mapped to proximate representations, and 2) how attention components capture the conceptual alignment within and across modalities. 
\end{itemize}

\section{Related Work}

As deep learning models flourish, many tools and methods have been proposed to offer insight into their inner workings. Some methods are general-purpose \cite{nori2019interpretml,wexler2019if}, while others are nuanced for specific models such as CNNs \cite{zhang2018interpreting,zhang2019interpreting} or RNNs \cite{hou2020learning, strobelt2017lstmvis, karpathy2015visualizing}. 
In transformers, the introduction of attention not only helped improve performance, but also served as an additional component towards interpretability. 

\textbf{Interpretability of NLP transformers} was initially approached through the analysis of attention to capture its alignments with syntactic or semantic relationships \cite{vig2019analyzing,Clark2019WhatDB}. 
Following this, subsequent works introduced additional functionalities including visualizations of hidden representations, task matching of attention heads, aggregate metrics, and interactive datapoint analysis \cite{tenney-etal-2020-language, hoover-etal-2020-exbert,lal-etal-2021-interpret,li-etal-2021-t3}. 
While common in allowing a user to understand the inner workings of transformers, each tool introduces novel applications.
For example, LIT \cite{tenney-etal-2020-language} enables probing for bias through examination of coreferences in counterfactuals. 
InterpreT \cite{lal-etal-2021-interpret} allows tracking of token representations through the layers and offers users the ability to define new metrics to identify coreference relationships in attention heads.
Additionally, T\textsuperscript{3}-Vis \cite{li-etal-2021-t3} focused on allowing users to improve transformer training by integrating the training dynamics in their visualization tool. 

\textbf{Interpreting vision transformers}, such as those for object detection \cite{carion2020end, dosovitskiy2021an} or image captioning \cite{cornia2020meshed,li2019entangled}, has also become increasingly popular.
Cordonnier \etal \cite{cordonnier2020on} showed that the first few layers in transformers can learn to behave similarly to convolutional layers, and demonstrated the filter patterns through visualization of image-to-image attention. 
As illustrated in this paper, the attention mechanism in transformers is a natural gateway to understanding vision models, as the heatmaps of attention can be used to highlight salient image regions. 
Furthermore, Chefer \etal \cite{chefer2021transformer} proposed a method for visualizing self-attention models by calculating a LRP \cite{bach2015pixel}-based relevancy score for each attention head in each layer, and propagating relevancies through the network. The end result is a class-specific visualization of image regions that led to the classification outcome.

\textbf{Multimodal interpretability} has, up until now, mainly entailed using probing tasks to study the impact of each modality on the responses generated by the model.
These probing tasks aim to quantify the information captured in the hidden representations by training classifiers, or applying metrics to embeddings at different points in a model. 
For instance, Cao \etal \cite{cao2020behind} proposed various probing tasks to analyze VL transformers, where the authors observed modality importance during inference, and identified attention heads tailored for cross-modal interactions as well as alignments between image and text representations. 
Additionally, other works have proposed probing tasks to interpret VL transformers for aspects such as visual-semantics \cite{dahlgren-lindstrom-etal-2020-probing}, verb understanding \cite{hendricks-nematzadeh-2021-probing}, and other concepts such as shape and size \cite{salin2022vision}.
However, a disadvantage of probing tasks is the amount of work: additional training of the classifiers is often required, and specific task objectives must be defined to capture different embedded concepts. 
Finally, most aforementioned works require image-caption pairs as input, and are therefore not best suited for interpreting multimodal transformers in tasks such as visual question answering.

Recently, a first attempt at explaining predictions by a VL transformer was proposed in \cite{Chefer_2021_ICCV}.
There the authors constructed a relevancy map using the model's attention layers to track the interactions between modalities. 
The relevancy map is updated by a set of update rules that back-propagates relevancies of the prediction result back to the input. This map is very useful in understanding how model decisions are formed, but a more comprehensive interpretation for other aspects of VL transformers is still needed.

Our proposed tool, VL-InterpreT, differs from previous works in that it interprets various aspects of multimodal transformers in a single interface. 
This interactive interface allows users to explore interactions between tokens in each modality from a bottom-up perspective, without tying to task-specific inputs and outcomes. 
To the best of our knowledge, this is the first interactive tool for interpreting multimodal transformers. 

\section{System Design}
\subsection{Workflow}

\begin{figure*}
  \centering
  \includegraphics[width=\textwidth]{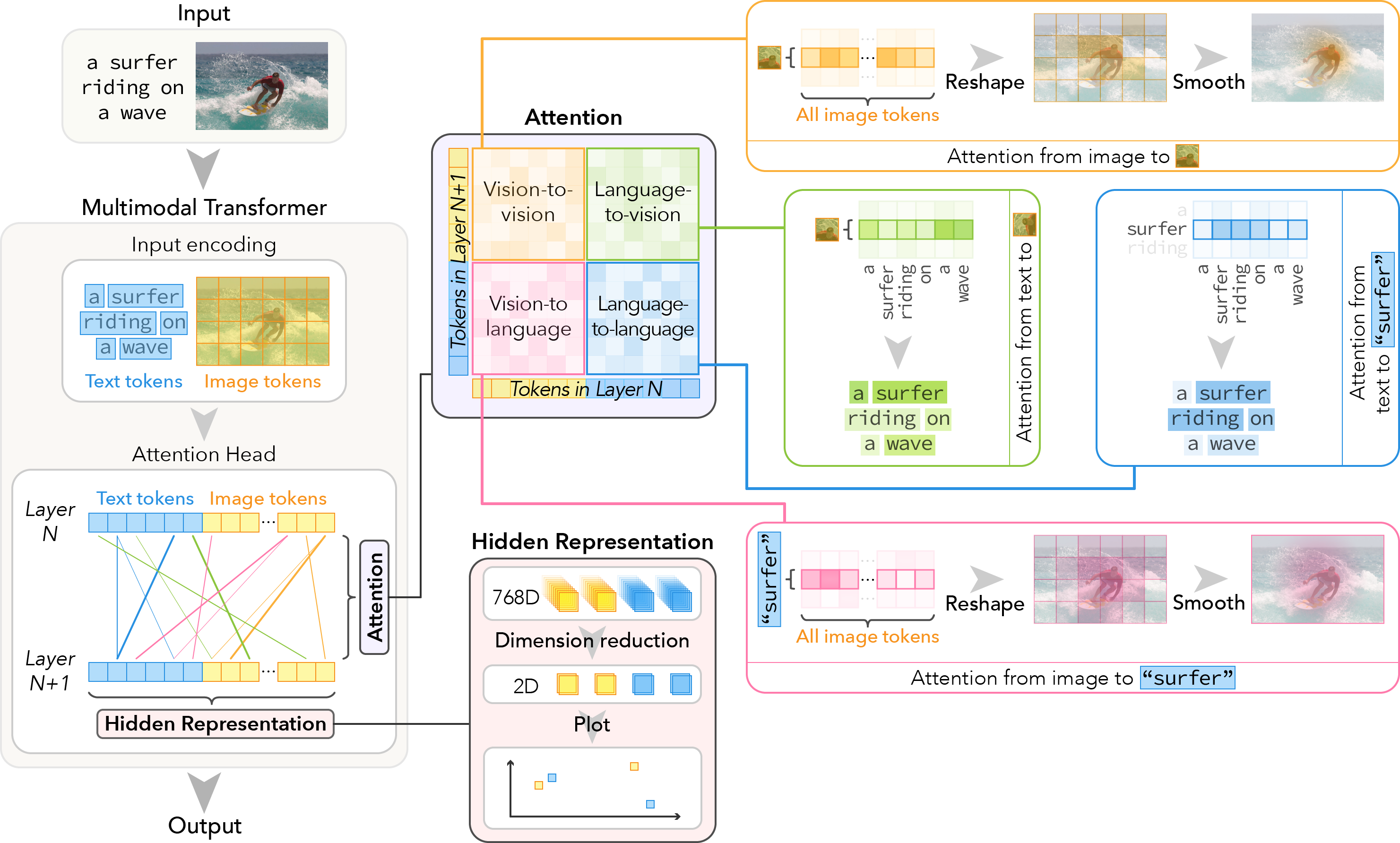}
  \caption{VL-InterpreT workflow. Data shown in this figure are for illustration purposes only.}
  \label{xmodal_flow}
  \vspace{-1em}
\end{figure*}

\begin{figure*}[h!]
  \centering
  \begin{minipage}[t]{.56\linewidth}
    \includegraphics[width=\textwidth]{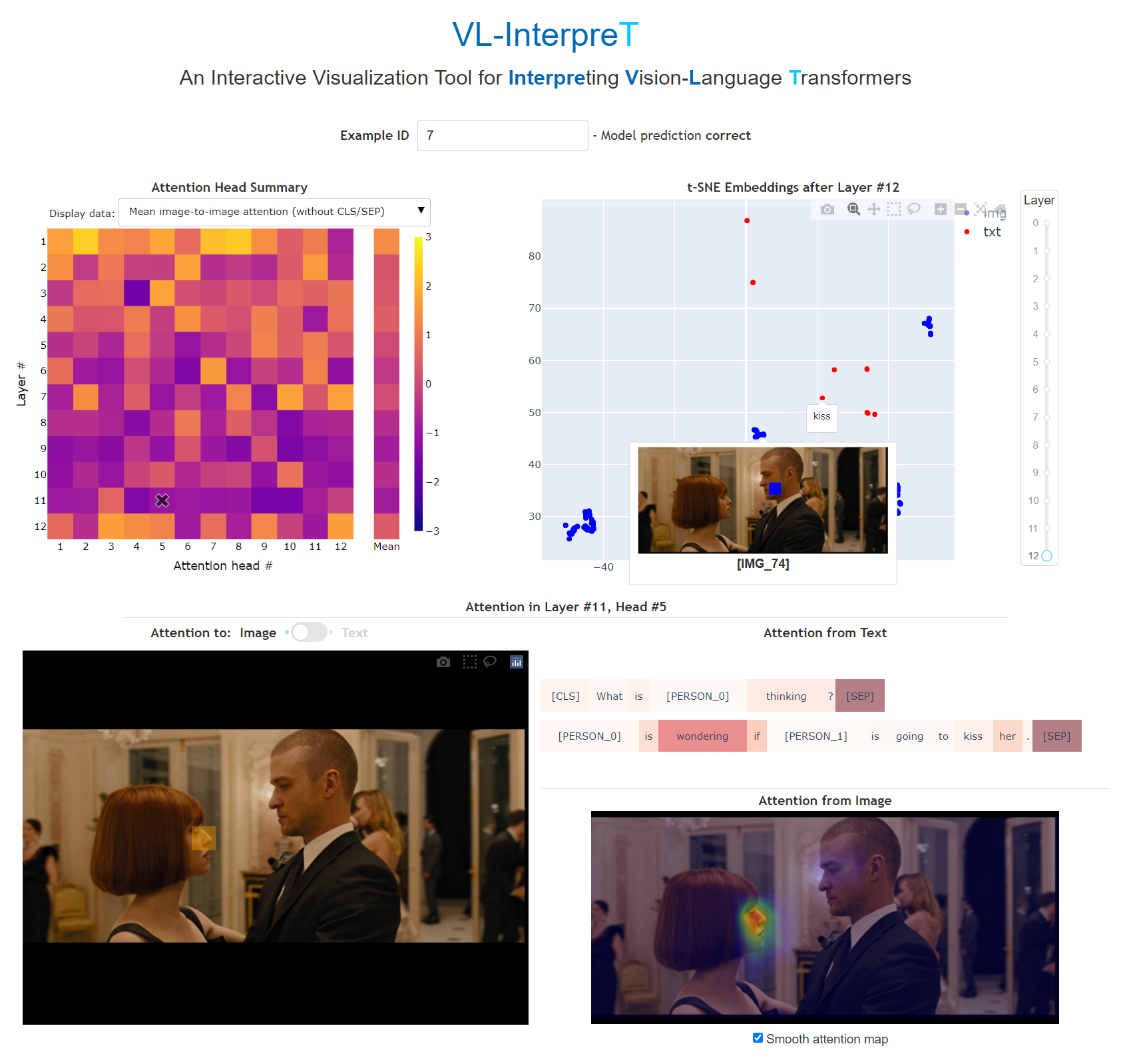}  
    \caption{The VL-InterpreT user interface (rearranged for print).} 
    \vspace{-1em}
    \label{app_ui}
 \end{minipage} 
 \hfill
 \begin{minipage}[t]{.36\linewidth}
    \centering
    \includegraphics[width=\textwidth]{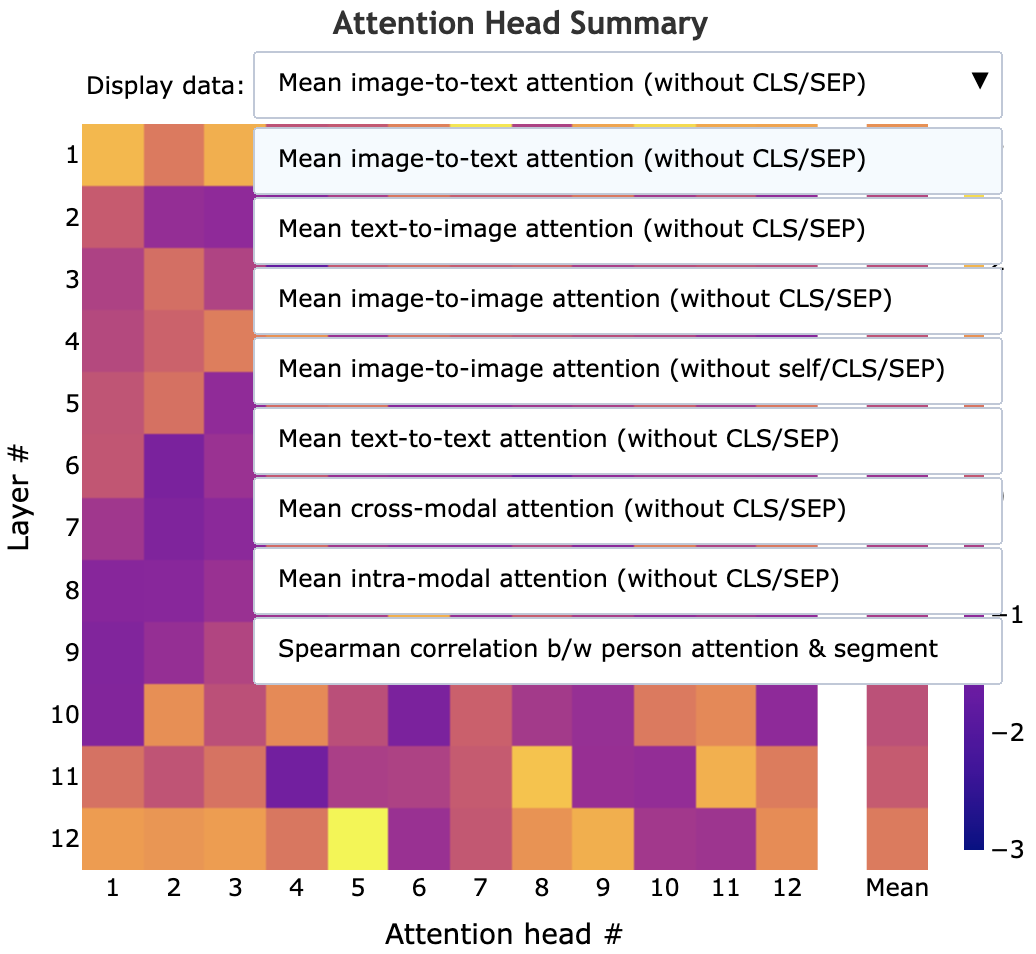}
    \caption{The Attention Head Summary plot colored by the metrics selected from the dropdown menu}
    \label{fig:drop_down_menu_stats}
 \end{minipage} 
\end{figure*}

VL-InterpreT is designed as a two-stage workflow: First, the attentions and hidden states of a given multimodal model are generated and saved for a set of examples (in this case, 100 examples). 
Next, the saved data, along with the metadata of the corresponding examples, are loaded into our tool to enable visualizations of the inner workings of the model. The workflow of VL-InterpreT is shown in \Cref{xmodal_flow}, and the user interface is shown in \Cref{app_ui}.
Different from interpretability tools for NLP transformers, VL-InterpreT addresses the analysis and interpretation of the following properties of multimodal transformers:

\begin{itemize}

    \item \textbf{Input:}~~
    In general, multimodal transformers are able to process inputs originating from different modalities, \eg, video, audio, or language.
    Here, we only consider VL transformers where the input is composed of visual and textual tokens. 
    These tokens are mapped into a shared space, allowing for concept-level alignment between the two modalities. 
    
    \item \textbf{Attention:}~~
    Because of the bi-modal nature of the input, the resulting attention can be split into four components: language-to-language, vision-to-vision, vision-to-language, and language-to-vision. An illustration of these components is shown in Figure \ref{xmodal_flow}. 
    
    \item \textbf{Hidden states:}~~
    In each layer, the transformer produces as many hidden states as the number of input tokens. Each input token is embedded as a \emph{d}-dimension vector after processed by each layer.
    
  \end{itemize}

\subsection{Visualizations}

\subsubsection{Attention heads components}
\label{sec:system_design_viz_attn}
In this section, we describe each attention component and how VL-InterpreT visualizes them interactively.
The attention matrix in a VL transformer, as they are loaded in VL-InterpreT, is of size $(N_{layers}, N_{heads}, L_{v}+L_{l}, L_{v}+L_{l})$, where $N_{layers}$ and $N_{heads}$ correspond to the number of layers and heads, respectively, $L_v$ to the number of visual tokens, and $L_l$ to the number of text tokens.
\begin{itemize}
    \item The \textbf{Language-to-Vision} attention component (L2V) of size $(N_{layers}, N_{heads}, L_{l},  L_{v})$ reflects the text tokens' dependency on the visual tokens.
    This partition of the attention contains the attention scores calculated from the dot product of the query vector based on the selected image patch and the key vectors from text tokens.
    These attention scores are the weights given to value vectors of text tokens when summing for the updated representation of a specific image patch in the next layer. 
    A user can select any attention head and image patch in the interface of our tool, and the corresponding L2V attention weights are displayed as a heatmap overlaid on the input text, to visualize how much each text token contributes to the updated image patch representation (see \Cref{x2v_ex}).

    \item The \textbf{Vision-to-Language} attention component (V2L) of size $(N_{layers}, N_{heads}, L_{v},  L_{l})$ reflects the visual tokens' dependency on the text tokens.
    This component of attention in a VL transformer arises through the query-key dot product where the query vectors are computed from text token embeddings, and the key vectors are computed from the image token embeddings.
    A user can select a specific attention head and a text token, and the corresponding attention scores will be overlaid onto the image as a heatmap (see Figure \ref{x2l_ex}). 
    This visualization helps users understand the relative contributions of various parts of the image to the updated representations of the text tokens.
    The interactive application also allows users to play an animation, in which the heatmap over the image is automatically displayed in a sequence for each word.

    \item The \textbf{Language-to-Language} attention component (L2L) of size $(N_{layers}, N_{heads}, L_{l},  L_{l})$ corresponds to the attention mapping in NLP Transformers. 
    This component visualizes how all text tokens attends to each individual token of the input sentence (see Figure \ref{x2l_ex}).
    
    \item The \textbf{Vision-to-Vision} attention component (V2V) of size $(N_{layers}, N_{heads}, L_{v},  L_{v})$ is analogous to language-to-language attention, but in the visual space. It represents the attention between one visual tokens and all visual tokens, including itself. 
    Similar to the V2L component, an attention vector (of size $(L_{v}, 1)$) here in a given head can also be translated into a heatmap and overlaid onto the image. 
    This visualization is useful for identifying the contributions of different image patches to the updated representation of the image patch selected by a user. 
    
    \end{itemize}

\subsubsection{Attention head summary}
\label{sec:attention_head_summary}
This functionality allows users to visualize a head summary plot containing statistics of the attentions calculated for all heads and layers.
For an attention matrix of size $(N_{layers}, N_{heads},  L_{v} + L_{l}, L_{v} + L_{l})$, the head summary computes statistical metrics over the last two dimensions, resulting in a plot of size $(N_{layers}, N_{heads})$.
\begin{itemize}
    \item The \textbf{mean} attention in an attention head is generated by calculating the average of the corresponding attention matrix, while some tokens can be excluded from this calculation. A user can select a summary metric from the list of options. Each metric can be restricted to different components of the attention. For example, instead of computing the mean of all input tokens regardless of modality, the calculation can be limited to the vision-to-language part or the vision-to-vision part of the attention. Additionally, users may also focus on both cross-modal components (i.e., V2L and L2V averaged) or both intra-modal ones (i.e., V2V and V2L averaged). See the drop down menu in Figure \ref{fig:drop_down_menu_stats}.

    \item \textbf{Custom metrics:}
    Based on users' interests, custom metrics can be integrated in this plot to show relevant attention heads.
    For example, we created a custom metric to look for the attention heads responsible for aligning the same person between vision and language modalities, based on the V2L component. This metric, labeled \textit{Spearman correlation b/w person attention and segment} (see Figure \ref{fig:drop_down_menu_stats}), is the Spearman correlation between the panoptic segmentation mask for a person in the image (generated by Maskformer model \cite{maskformer}), and the attention heatmap to the corresponding person token in the Vision-to-Language attention component. 

    This metric allows a user to identify attention heads that perform a function similar to panoptic segmentation for people (see \Cref{fig:seg_corr}).

\begin{figure}
    \centering
    \includegraphics[width=\columnwidth]{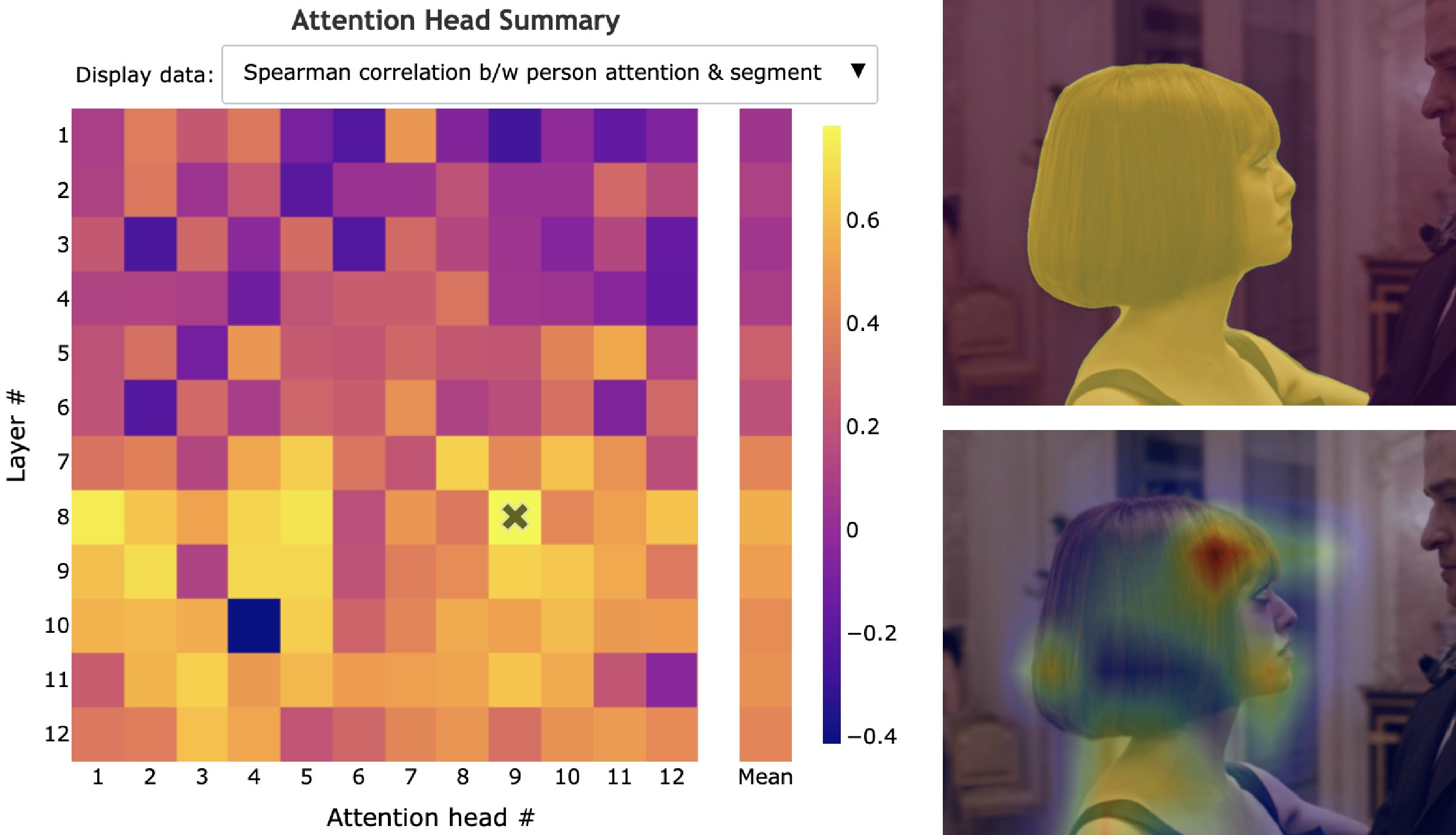}
    \caption{Custom metric: Average correlation between the V2L attention to the \token{PERSON} tokens (bottom right) and the person's panoptic segmentation mask (top right).}
    \vspace{-1em}
    \label{fig:seg_corr}
\end{figure}

\end{itemize}

Finally, for each metric, a mean is automatically computed for each layer and shown in the rightmost column of the attention head summary (see Figures \ref{app_ui} and \ref{fig:seg_corr}). This column visualizes the general behavior in each layer for the selected metrics, and shows its evolution throughout the layers of the transformer.

\subsubsection{Hidden state representation}
For each input token, a \emph{d-}dimensional hidden representation is produced by the transformer after each layer (in our setup, $d = 768$). 
This pool of hidden representations is then filtered by two criteria: (1) if the related text is a stop word and (2) if the related image patch comes from a part of the background (e.g., wall, ground, etc.).
In order to visualize the remaining hidden representations in a readable form, t-distributed Stochastic Neighbor Embedding (t-SNE) \cite{tsne} was applied to reduce dimensions and create disjoint t-SNE spaces for different layers. This way, given a selected example, VL-InterpreT tracks the hidden representations both before the first layer and after each subsequent layer, and plots them in two-dimensional spaces.\\

Figure \ref{fig:tsne_func} shows the data points representing the visual (in blue) and textual (in red) tokens from a given example. When hovering on a data point from language, the corresponding text is displayed. When hovering on a data point representing visual tokens, the image is shown with a highlighted blue patch corresponding to the visual token. Further observations on the hidden states often reveal the concept-level vision-text alignments that are learnt in this multimodal setup (see Section \ref{sec:vcr_hidden_states}). 
To help further understand this alignment, VL-InterpreT allows a user to select a token (text or image patch) from a given example, and shows the nearest token in the other modality from the whole subset of examples, marked with a green star.

\section{Case Studies}
To demonstrate the functionalities of VL-InterpreT, we analyze an end-to-end VL transformer model, KD-VLP \cite{liu2021kdvlp}, on two benchmarks: Visual Commonsense Reasoning (VCR) \cite{zellers2019recognition} and WebQA \cite{chang2021webqa}. Nonetheless, our tool is generally applicable to a variety of multimodal transformer configurations and types of VL datasets.

\subsection{Model}
The \textbf{KD-VLP model} used in our case study is a transformer for end-to-end vision-language processing. 
This model utilizes a ResNet backbone for visual inputs, and is pretrained using text-oriented, image-oriented, and cross-modal tasks in the form of masked language modeling, object-guided masked vision modeling, and phrase-region alignment, respectively. 
Depending on the application, the KD-VLP model can be fine-tuned for classification or generation tasks given bi-modal input of image and text. 

\subsection{Analysis on VCR}
\label{sec:vcr_analysis}
The \textbf{VCR benchmark} consists of 290K multiple choice QA problems derived from 110K movie scenes. This dataset is uniquely valuable in that it requires higher-order cognition and commonsense reasoning about the world. Given an image and a question, the objective is to select an appropriate answer from four possible choices, and then provide the rationale. 
To predict the correct answer and rationale, the KD-VLP model is fed with the image, the question, and each answer or rationale individually. The predicted answer $a_p$ is the answer/rationale choice that receives the highest probability score, \ie,
\begin{equation}
    a_p = \underset{a_j}{\arg\!\max}~f(v, q, a_j)
\end{equation}
where $f$ is the KD-VLP model, $v$ is the image, $q$ the question, and $\{a_j|~j \in [1,2,3,4]\}$ are the possible answer choices.

The functionalities of VL-InterpreT are demonstrated using example \textit{val-445} from the VCR validation set, or example ID 89 in the VL-InterpreT live demo. 
This data sample, shown in Figure \ref{x2l_ex}, comprises an image showing a little girl running to a man and a woman in a garden. The question is: 
\begin{center}
\textit{Where is \token{PERSON\_0} running to?}
\end{center}
where \token{PERSON\_0} corresponds to the little girl on the right side of the image (the locations of persons are provided in the VCR dataset and also passed to the model).
We also analyze the answer predicted by the model (which in this case is correct):

\begin{center}\textit{\token{PERSON\_0} is running to help \token{PERSON\_1} and \token{PERSON\_2} with the plants}.
\end{center}
where \token{PERSON\_1} and \token{PERSON\_2} refer to the couple.

The following analyses on this example will highlight the visualization capabilities that our application provides.

\subsubsection{Attention head summary}
\begin{figure}
    
  \centering
  \includegraphics[width=0.6\columnwidth]{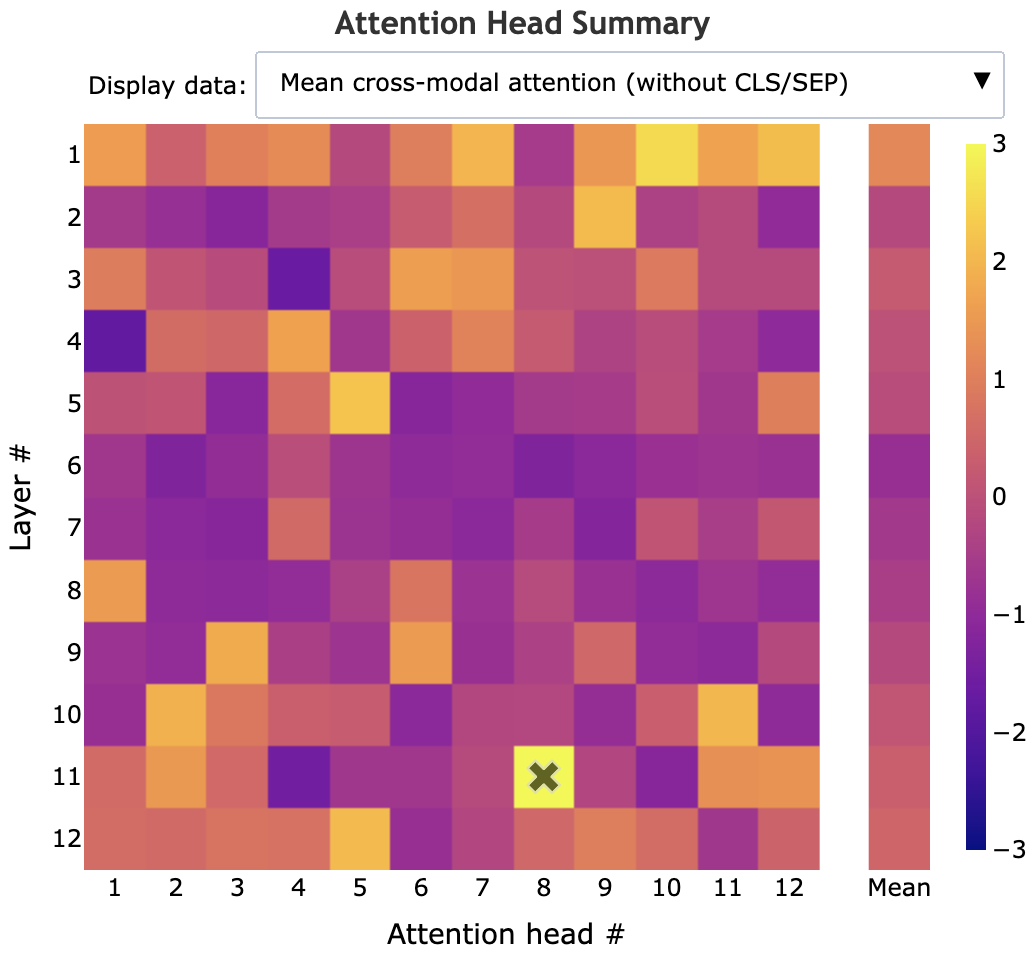}
  \caption{Attention Head Summary}
  \label{fig:stats}
  \vspace{-1em}
\end{figure}

This functionality allows a user to identify interesting heads based on various metrics.
By selecting \textit{Mean cross-modal attention} from the dropdown menu (see Figure \ref{fig:drop_down_menu_stats}), a user can identify attention heads specialized in cross-modal attention. For instance, in Figure \ref{fig:stats} the eighth attention head in layer 11 (denoted as (11, 8)) has, on average, the highest attention across modalities.
Thus, we focus on this specific head and plot its cross-modal (V2L and L2V) attention. Apart from cross-modal attention, specific heads could also been identified through \textit{Mean image-to-text attention} for analyzing V2L attentions, or \textit{Mean text-to-image attention} for L2V attentions.
Analogous procedure also applies to L2L and V2V attention components -- for instance, the metric \textit{Mean image-to-image attention (without self)} shows the heads' attentions from every image patch to the every other image patches, excluding each patch itself and its neighbors.
As described in section \ref{sec:attention_head_summary}, a custom metric was used to identify attention heads with V2L components for \token{PERSON} tokens highly correlated with their panoptic segmentations. As shown in Figure \ref{fig:seg_corr}, head (8, 9) scores particularly high in this metric. With head (8, 9) selected, the V2L attention attention heatmap for the person token (\Cref{fig:seg_corr}, bottom right) indeed aligns well with the panoptic segmentation of the corresponding woman (\Cref{fig:seg_corr}, top right). Furthermore, this correlation metric exhibits a trend where middle and later layers tend to have higher correlation coefficients on average (above 0.5) than early layers (less than 0.1), showing the evolution of attention patterns as the transformer layers grow deeper.

\subsubsection{Attention components}
\label{sec:vcr_attention_compo}
\begin{figure}
    \begin{subfigure}{\columnwidth}
      \centering
      \includegraphics[width=\linewidth]{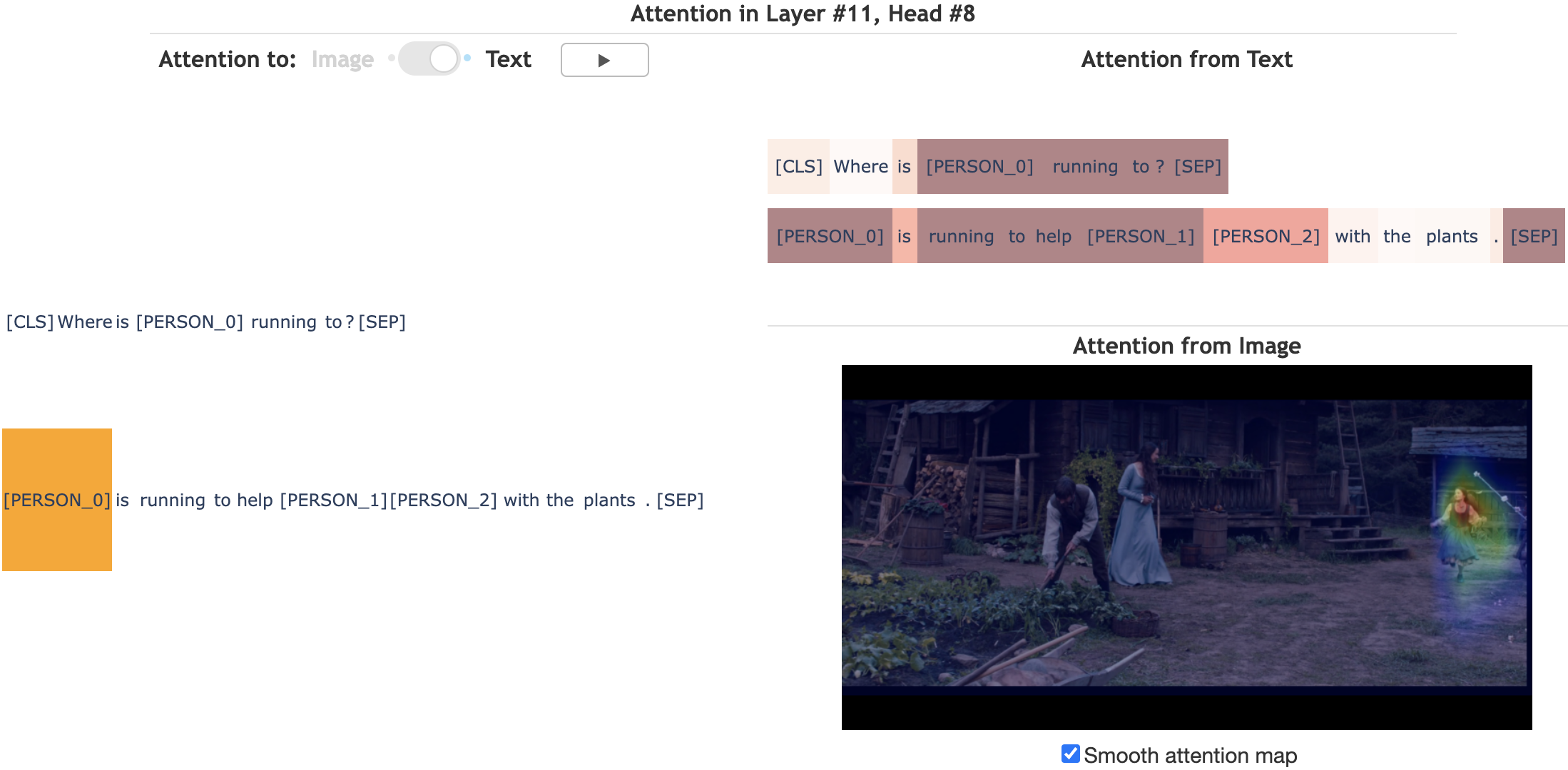}
      \caption{Attention to the text token \token{PERSON\_0}}
      \label{fig:x2l_ex1}
    \end{subfigure}
    \begin{subfigure}{\columnwidth}
             \centering
      \includegraphics[width=\linewidth]{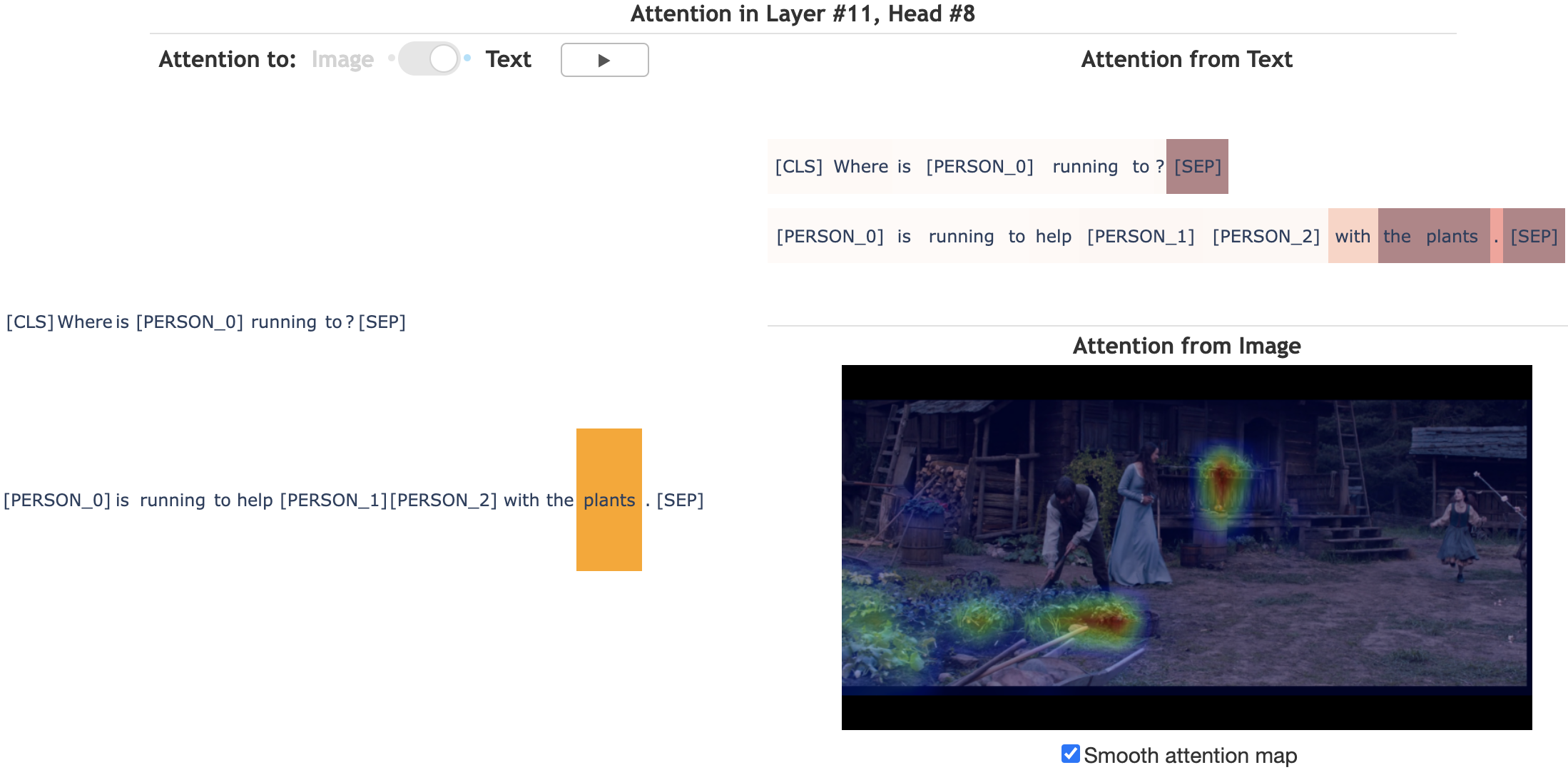}
      \caption{Attention to the text token \textit{plants}}
      \label{fig:x2l_ex2}
    \end{subfigure}

    \caption{Two selected text tokens and the corresponding L2L (top right) and V2L (bottom right) attention to them.}
    \label{x2l_ex}

\end{figure}

As described in \Cref{sec:system_design_viz_attn}, \Cref{x2l_ex} shows the heatmaps over the image and the text generated by the \textbf{Language-to-Language} and \textbf{Vision-to-Language} attention components of head (11, 8). This figure shows how attentions differ for two selected tokens: \token{PERSON\_0} and \textit{plants}. 
The V2L components are represented as heatmaps over the images at the bottom right of Figures \ref{fig:x2l_ex1} and \ref{fig:x2l_ex2}. It can be observed that the attention is concentrated on regions corresponding to the text, namely the little girl for the attention to \token{PERSON\_0} (Figure \ref{fig:x2l_ex1}) and the plants for the attention to the \textit{plants} token (Figure \ref{fig:x2l_ex2}).
The L2L components are on the top right of these figures. For both selected text tokens, related text tokens (including themselves) are highlighted in the heatmaps. For the example in Figure \ref{fig:x2l_ex1}, the attention to \token{PERSON\_0} is mostly from \textit{\token{PERSON\_0}},  \textit{running to}, and \textit{running to help \token{PERSON\_1}}. In the other example in Figure \ref{fig:x2l_ex2}, the attention to \textit{plants} is mostly from the \textit{plants} token itself.

\begin{figure}[tb]
      % include second image
      \includegraphics[width=0.9\linewidth]{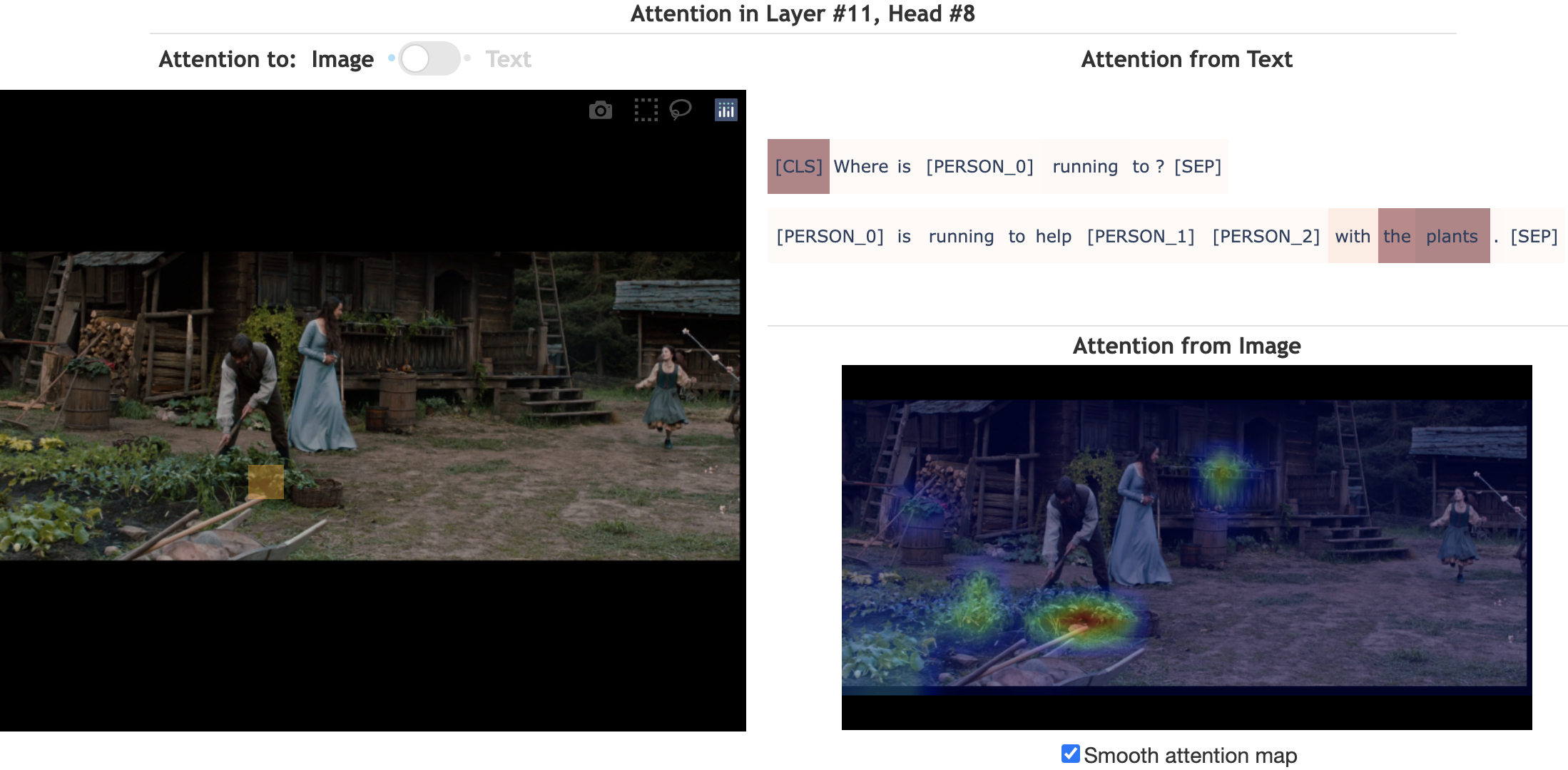}  

\caption{Attentions to vision for a selected image patch in the left image. The heatmaps show the attention from the text (top right) as well as other patches in the image (bottom right).}
\label{x2v_ex}
\vspace{-1em}
\end{figure}

Figure \ref{x2v_ex} visualizes the attentions to vision tokens, i.e., the \textbf{Language-to-Vision} and the \textbf{Vision-to-Vision} attentions.
    In this example, we select an image patch that is a part of the plants on the left image for analysis. Accordingly, the L2V component (top right of \Cref{x2v_ex}) shows that the attention to this image patch is mainly from the \token{CLS} token as well as the \textit{plants} token, which aligns with the concept behind the selected region. As for the V2V component (bottom right of the figure), it is also interesting that most of the regions containing plants in the image attend to this specific patch of plants, which again shows a conceptual alignment.
In summary, this example shows evidence of a unified concept of ``plants", where the attention has a consistent pattern between intra- and cross-modal components.

\subsubsection{Hidden states}
\label{sec:vcr_hidden_states}
\begin{figure}
        \begin{subfigure}{\columnwidth}
          \centering
          % include second image
          \includegraphics[width=0.8\linewidth]{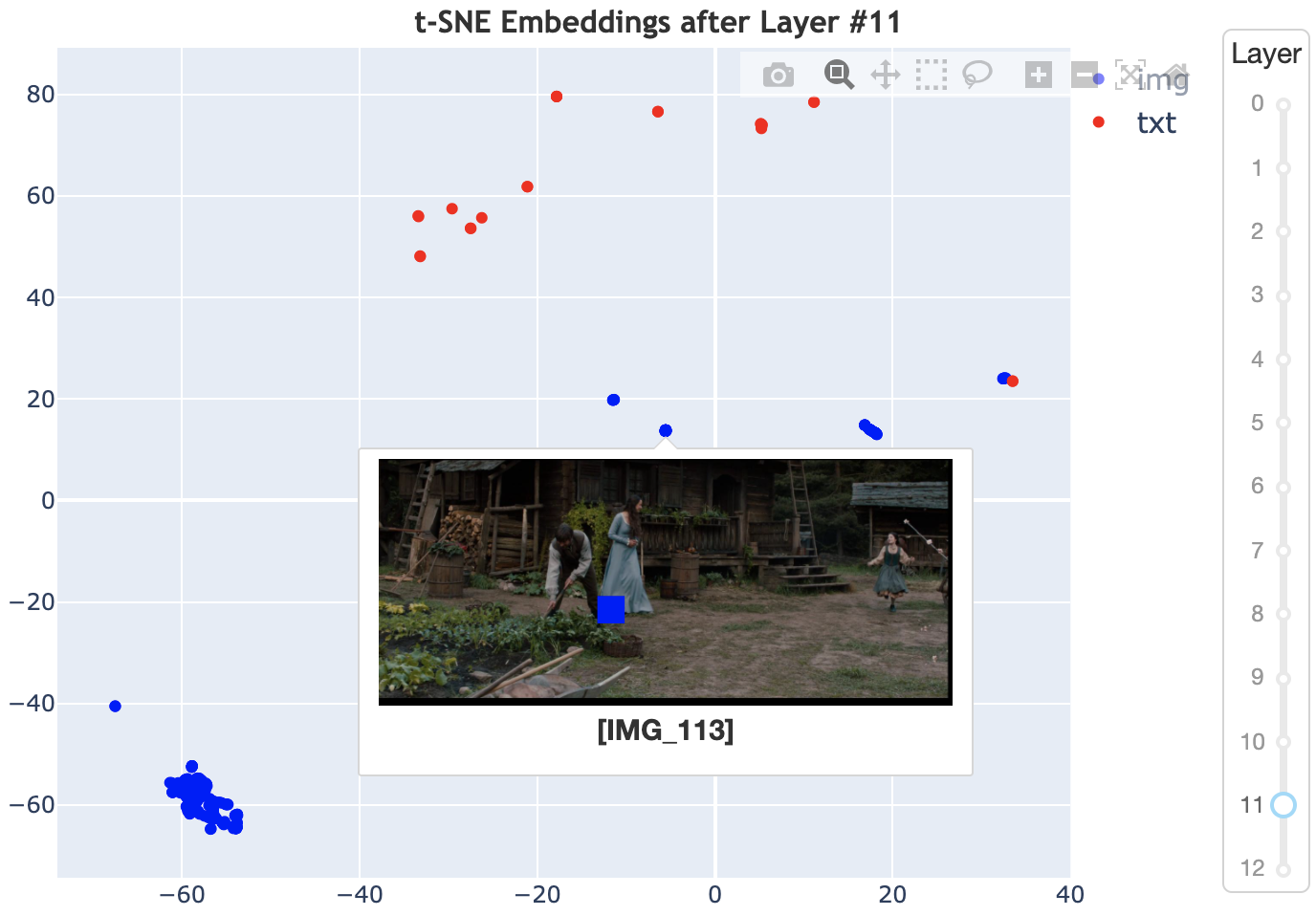}  
          \caption{Hovering over a data point representing a visual token displays the corresponding image patch.}
          \label{fig:tsne_func}
        \end{subfigure}
        % \hfill
        \begin{subfigure}{\columnwidth}
          \centering
          % include third image
          \includegraphics[width=0.85\linewidth]{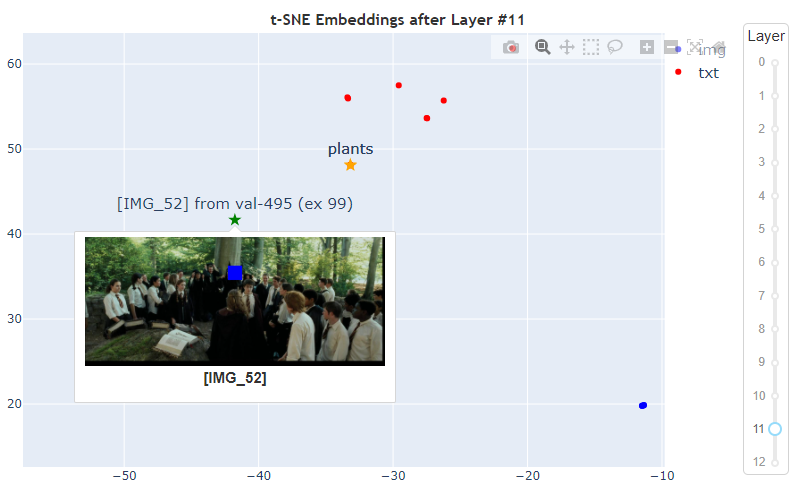}  
          \caption{Clicking on the text token \textit{plants} marks it orange, and displays the closest image token from the whole dataset, marked as a green star.}
          \label{fig:tsne_plants}
        \end{subfigure}
        \caption{t-SNE plot from the hidden representations of the selected example.
        }
        \label{fig:tsne}
        \vspace{-1em}
    \end{figure}

Figure \ref{fig:tsne} demonstrates the capabilities of the VL-InterpreT for visualizing hidden states. When selecting the text token \textit{plants} (marked in orange) from the previous example \textit{val-445} (ex 89), Figure \ref{fig:tsne_plants} shows that in layer 11, the closest image patch from the whole pool of examples is the \token{IMG\_52} (marked with a green star) from \textit{val-495} (ex 99). It is interesting that even when it comes from a very different example, this token also refers to the plants in the image. Other than layer 11, users can select different layers on the right to see other closest image tokens to \textit{plants} throughout layers.

\begin{figure*}[htp]
    \centering
    \begin{minipage}{.55\linewidth}
        \begin{subfigure}[t]{.46\linewidth}
          \includegraphics[width=\linewidth]{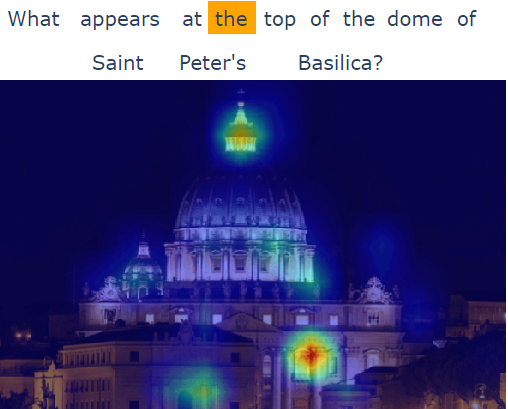}  
          \caption{\textbf{Predicted}: \textit{A cross appears at the top of the dome of Saint Peter's Basilica.}}
          \label{fig:correct_webqa_cross}
        \end{subfigure} 
        \hfill
        \begin{subfigure}[t]{.46\linewidth}
            \includegraphics[width=\linewidth]{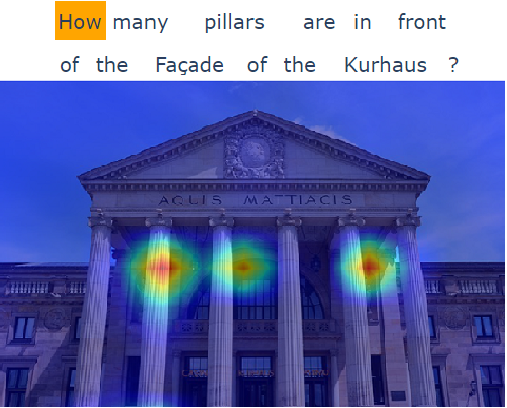}  
            \caption{
                \textbf{Predicted}: \textit{There are 6 pillars in front of the Façade of the Kurhaus}.
            }
          \label{fig:correct_webqa_columns}
        \end{subfigure}
        \begin{subfigure}[t]{.46\textwidth}
          \includegraphics[width=\linewidth]{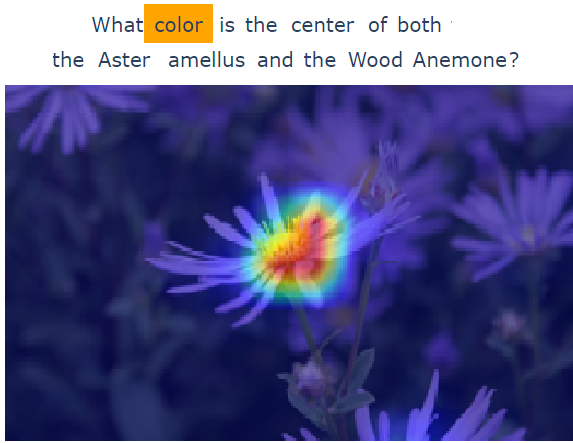}  
          \caption{
        \textbf{Predicted}: \textit{The center of both the Aster amellus and the Wood Anemone is yellow.}
         }
          \label{fig:correct_webqa_flower}
        \end{subfigure}
        \hfill
        \begin{subfigure}[t]{.46\textwidth}
          \includegraphics[width=\linewidth]{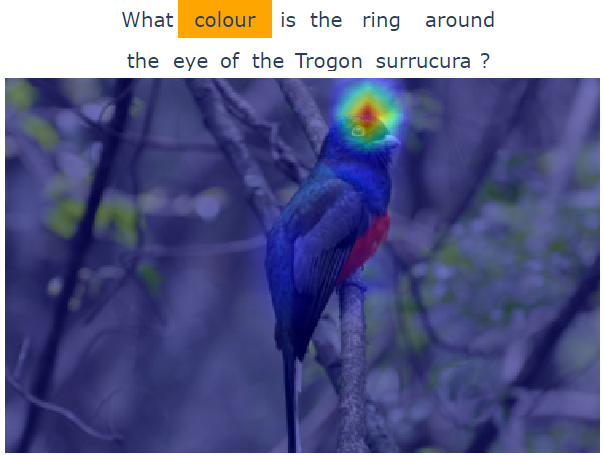}  
          \caption{
        \textbf{Predicted}: \textit{The ring around the eye of Trogon surrucura is red.}
        }
          \label{fig:correct_webqa_bird}
        \end{subfigure}

    \end{minipage}
    \hfill
    \begin{minipage}{.42\linewidth}
        \begin{subfigure}[t]{.48\linewidth}
            \includegraphics[width=\textwidth]{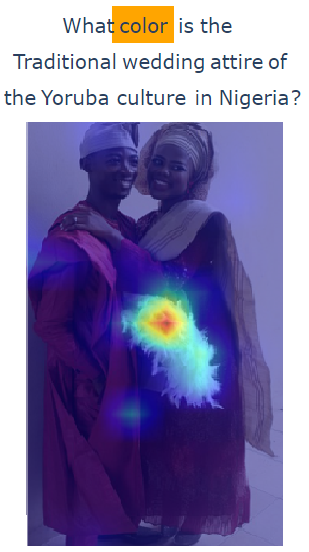}
            \caption{
                \textbf{Predicted}: \textit{The Traditional wedding attire of the Yoruba culture in Nigeria is \textbf{white}.} \\
                \textbf{Correct color}: \textbf{\textit{maroon}}.
                }
            \label{fig:incorrect_webqa_wedding}
        \end{subfigure}
        \hfill
        \begin{subfigure}[t]{.45\linewidth}
            \includegraphics[width=\linewidth]{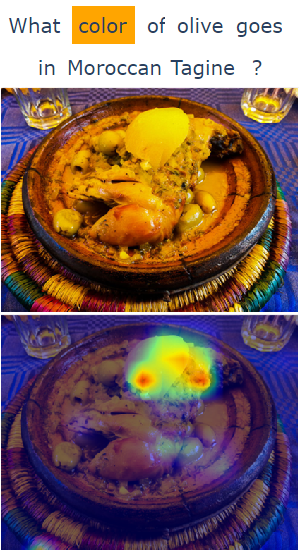}  
            \caption{
                \textbf{Predicted}: \textit{The olive that goes in Moroccan Tagine is \textbf{yellow}.} \\
                \textbf{Correct color}: \textbf{\textit{green}}.
                 }
            \label{fig:incorrect_webqa_olive}
        \end{subfigure}
    \end{minipage}
    \caption{V2L attention heatmaps for WebQA examples and generated answers. Figures have been rearranged for print. (a)-(d) are correct predictions and (e) and (f) are incorrect.}
    \label{fig:webqa_examples}
    \vspace{-1em}
\end{figure*}

Sections \ref{sec:vcr_attention_compo} and \ref{sec:vcr_hidden_states} show evidence of alignment between visual and textual concepts of \textit{plants}.
We see that the concept of "plants" in both modalities and across examples is captured by proximate representations.
As such, VL-InterpreT allows studying how such sense of objectness emerges by probing the attentions and hidden states across all layers of the transformer.
\subsection{Analysis on WebQA}

\label{sec:webqa_analysis}

The \textbf{WebQA benchmark} focuses on multimodal, multihop reasoning for open-domain question answering. 
This benchmark emulates a knowledge-seeking query to a search engine for information which may be contained in either text-based articles or images. 
Given a query, the goal is to identify which information is relevant across modalities, and to generate a full natural language answer based on the selected sources. 
The dataset contains 50k QA pairs, half of which are text-based and the other half are image-based. 

We use the KD-VLP model to first select relevant sources using a classification head. Then, by adding a decoder in a Fusion-in-Decoder manner as in \cite{yu2022kgfid}, a predicted answer is generated based on the retrieved sources. 
Similar behaviors can be studied for WebQA as in the previous section. The following analyses will focus on the L2V attention in the KD-VLP encoder, as these visualizations helps in understanding how the model generates answers. 

For this analysis, attention head (11, 5), identified in the same way as previous analyses, was selected.
Figure \ref{fig:webqa_examples} shows the V2L attention from image to the highlighted word above the pictures. These heatmaps show that the model attends more to the regions that help answer the question. For instance, when the question is about pillars, Figure \ref{fig:correct_webqa_columns} shows that the model attention comes particularly from the columns in the picture.
In order to determine the color of the center of the flower in Figure \ref{fig:correct_webqa_flower}, the model exhibits attention from the flower center in the image and generates the correct color (yellow). 
Furthermore, these visualizations can also be generated for incorrectly answered questions, providing insights into the reason why incorrect answers were generated. For example, one may imagine why the model answered incorrectly in Figure \ref{fig:incorrect_webqa_wedding} and \ref{fig:incorrect_webqa_olive}: In \ref{fig:incorrect_webqa_wedding}, the attire that the question asks about was misidentified. That is, instead of getting attention from patches of the maroon dress, the model focuses on the white feathers. A similar behavior is also seen in \ref{fig:incorrect_webqa_olive}, where the model fails to locate the olives but focuses on the yellow vegetable in the tagine. In both cases, the model answers according to the identified object color (i.e., "white" feathers and "yellow" olive).
These interpretations of attention helps users identify why a model fails in certain cases, and provides guidance for future efforts in improving model accuracy.

\section{Conclusions and Future Directions}
In this paper we presented VL-InterpreT, an interactive visualization tool for interpreting vision-language transformers. 
This tool allows for interactive analysis of attention and hidden representations in each layer of any VL transformer.
VL-InterpreT can be used to freely explore the interactions between and within different modalities to better understand the inner mechanisms of a transformer model, and to obtain insight into why certain predictions are made.
Through case studies, we demonstrated how VL-InterpreT can be used to validate the learning of cross-modal concepts, as well as to ``explain" cases of failure.

In the latest version, VL-Interpret is able to run a live model to process user-generated examples in real time.
This allows interactive manipulations of inputs, including both text and image, to study their effects on the attention and hidden representations. 

For future work, we would like to include aggregated metrics and visualizations over multiple samples to obtain a more comprehensive understanding of model operation.
In addition, we hope to experiment with any additional functionalities that will assist users in interpreting multimodal transformers, and continue to enhance this interpretability tool.

%%%%%%%%% REFERENCES
% \clearpage
{\small
\bibliographystyle{ieee_fullname}
\bibliography{egbib}
}

\end{document}